\documentclass[runningheads]{llncs}
\usepackage[T1]{fontenc}
\usepackage{graphicx,verbatim}
\usepackage{multirow}
\usepackage{amsmath} 
\usepackage{wrapfig}
\usepackage{amssymb}
\usepackage{amsfonts}
\usepackage{booktabs}
\usepackage{color}
\usepackage{graphicx}
\usepackage{caption}
\usepackage{subcaption}
\usepackage{bbm}
\usepackage{bm}
\usepackage{xcolor}
\usepackage{tikz}
\usepackage{etoolbox}
\usepackage{colortbl}
\usepackage{tablefootnote}
\usepackage{makecell}
\usepackage{mwe} 
\usepackage{hyperref} 
\hypersetup{
    colorlinks=true,     
    linkcolor=black,     
    citecolor=black,     
    filecolor=black,     
    urlcolor=blue        
}

\newrobustcmd*{\mycircle}[1]{\tikz{\filldraw[draw=#1,fill=#1] (0,0) circle [radius=0.1cm];}}

\newrobustcmd*{\mytriangle}[1]{\tikz{\filldraw[draw=#1,fill=#1] (0,0) --
(0.2cm,0) -- (0.1cm,0.2cm);}}

\definecolor{cvprblue}{rgb}{0.21,0.49,0.74}
\definecolor{darkyellow}{rgb}{0.85, 0.65, 0.13}

{
\begin{document}
\title{PTCMIL: Multiple Instance Learning via Prompt Token Clustering for Whole Slide Image Analysis}
\titlerunning{PTCMIL}

\author{
Beidi Zhao\inst{1,2}\textsuperscript{*} \and 
SangMook Kim\inst{3}\textsuperscript{*\dag} \and
Hao Chen\inst{4} \and
Chen Zhou\inst{1,5} \and
Zu-hua Gao\inst{1,5} \and
Gang Wang\inst{1,5} \textsuperscript{\ddag} \and
Xiaoxiao Li\inst{1,2} \textsuperscript{\ddag}
}
\authorrunning{Zhao et al.}
\institute{%
The University of British Columbia \and Vector Institute \and Chungnam National University \and The Hong Kong University of Science and Technology \and BC Cancer Agency\\
\email{beidiz@student.ubc.ca, sangmook.kim@cnu.ac.kr, jhc@cse.ust.hk, czhou@bccancer.bc.ca, zuhua.gao@ubc.ca, gang.wang1@bccancer.bc.ca, xiaoxiao.li@ece.ubc.ca}
}

\maketitle           
\footnotetext[1]{Equal contribution.}
\footnotetext[4]{Work done while the author was a postdoctoral fellow at the University of British Columbia.}
\footnotetext[5]{Co-corresponding authors.}
\begin{abstract}
Multiple Instance Learning (MIL) has advanced WSI analysis but struggles with the complexity and heterogeneity of WSIs. Existing MIL methods face challenges in aggregating diverse patch information into robust WSI representations. While ViTs and clustering-based approaches show promise, they are computationally intensive and fail to capture task-specific and slide-specific variability. To address these limitations, we propose PTCMIL, a novel Prompt Token Clustering-based ViT for MIL aggregation. By introducing learnable prompt tokens into the ViT backbone, PTCMIL unifies clustering and prediction tasks in an end-to-end manner. It dynamically aligns clustering with downstream tasks, using projection-based clustering tailored to each WSI, reducing complexity while preserving patch heterogeneity. Through token merging and prototype-based pooling, PTCMIL efficiently captures task-relevant patterns. Extensive experiments on eight datasets demonstrate its superior performance in classification and survival analysis tasks, outperforming state-of-the-art methods. Systematic ablation studies confirm its robustness and strong interpretability. The code is released at 
\url{https://github.com/ubc-tea/PTCMIL}.

\end{abstract}

\section{Introduction}
\label{sec:intro}
Histopathology is the gold standard for cancer diagnosis, essential for tumor detection, subtyping, and survival prediction. With advancements in deep learning, digital pathology, which analyzes whole slide images (WSIs), has gained prominence \cite{bejnordi2017diagnostic,bulten2022artificial,cancer2013cancer}. WSIs are massive giga-pixel images that require computationally efficient processing, typically through multiple instance learning (MIL), which enables slide-level annotation without patch-level labels. However, WSIs exhibit significant inherent heterogeneity, containing diverse cell types and tissue structures with varying morphological and staining characteristics \cite{chan2023histopathology,hou2022h}. A key challenge in MIL is aggregating redundant patch information into robust WSI representations. Early MIL approaches treated patches independently, neglecting interactions, but recent methods leverage patch relationships for improved modeling \cite{shao2021transmil,chan2023histopathology}. The introduction of Vision Transformers (ViTs) \cite{shao2021transmil,xiang2023exploring} has enhanced global dependency modeling through self-attention.
Despite their effectiveness, ViTs face computational bottlenecks and overfitting issues, limiting their scalability in MIL applications \cite{yang2024mambamil}.

To address above issues and handle the vast diversity of patches within each WSI, recent methods incorporate clustering techniques to identify representative prototypes, thereby enhancing WSI representations \cite{yan2022deep,song2024morphological}. These approaches typically follow a two-stage process~\cite{song2024morphological}: (1) unsupervised clustering groups patches into prototypes, often leveraging global clustering across all patches, and (2) a pooling model trains on these prototypes for prediction over each WSI. While effective in capturing shared patterns, these methods face key limitations: (i) standalone clustering is not optimized for downstream tasks, potentially missing task-specific features, (ii) global clustering is computationally expensive, requiring patch sampling that may omit critical regions, and (iii) uniform centroids across WSIs fail to account for slide-specific variability, reducing adaptability. These challenges lead to our research question: \emph{How can we optimize patch clustering alongside WSI-level analysis efficiently and effectively?}

Visual Prompting (VP) \cite{jia2022visual}, adapted from natural language processing, enables ViTs to focus on specific tasks without extensive retraining. Prior work \cite{wang2023review} highlights that learnable prompt tokens enhance flexibility and efficiency across visual tasks. To address our research question, we propose \textbf{PTCMIL}, a \underline{P}rompt \underline{T}oken \underline{C}lustering-based ViT for \underline{MIL} aggregation, integrating clustering, prototyping, and downstream tasks in an end-to-end manner. Unlike traditional two-stage clustering methods, PTCMIL introduces prompt tokens to dynamically guide task-relevant clustering, capturing WSI patch heterogeneity while optimizing WSI-level analysis. We introduce projection-based clustering tailored to each WSI, reducing complexity compared to global clustering \cite{song2024morphological} while adding minimal parameters. PTCMIL improves prototype representations and WSI prediction. Our contributions are: 
(1) an end-to-end framework dynamically aligning clustering for prototyping with WSI-level analysis to enhance feature relevance for downstream tasks and improving performance and interpretability,
(2) efficient token clustering using prompt tokens and projection-based merging tailored to each WSI, preserving heterogeneity while reducing complexity, and
(3) extensive experiments demonstrating effectiveness across multiple WSI tasks, with systematic ablations validating design robustness.

\section{Methodology}
\label{sec:method}
This section introduces the overall pipeline of our method, which can simultaneously learn downstream task-related prototypes of WSI during the training (Fig.~\ref{fig:main}). The architecture of our PTCMIL consists of three parts: 1) Learnable prompt token-based clustering; 2) Prototype merging over clusters; 3) Pooling over prototypes to get the WSI representation for different downstream tasks. 
\subsection{Learnable Prompt Token-based Clustering}
\label{sec:clustering}
\begin{figure*}[t!]
  \centering
  \centerline{\includegraphics[width=\textwidth]{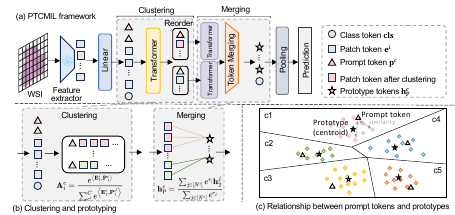}}
\caption{Overview of PTCMIL. (a) Overall framework, with patch feature tokens, prompt tokens and class token as input, and the objective prediction goal as output; (b) Interpretation of clustering and prototyping based on the clusters; (c) Interpretation of the relationship between prototypes and prompt tokens. 
}
\label{fig:main}
\end{figure*}

PTCMIL builds on the ViT-based MIL aggregation.
The introduced learnable prompts (
\tikz \fill[black](0,0) -- (1ex, 2ex) -- (2ex, 0) --cycle; in Fig.~\ref{fig:main}) are appended alongside patch tokens (
patch feature and represented as \tikz \fill[black] (0,0) rectangle (2ex, 2ex); in Fig.~\ref{fig:main}) and class tokens (represented as \tikz \fill[black](-2,0) circle (1ex); in Fig.~\ref{fig:main}) as the inputs of the ViT base model\footnote{We keep the class token following general VPT design~\cite{jia2022visual}.}. Denote $N$ as the number of patches of the given WSI. Denote $C$ as the desired number of clusters, a hyperparameters in our pipeline, and we have $C \ll N$. We represent the token embeddings of prompts, patches, and class tokens after the linear embedding layer (\textcolor{blue}{blue} block in Fig.~\ref{fig:main}) as $\mathbf{P}_0 = [\mathbf{p}_0^1, \cdots, \mathbf{p}_0^C] \in \mathbb{R}^{C \times D}$, $\mathbf{E}_0 = [\mathbf{e}_0^1, \cdots, \mathbf{e}_0^N]\in \mathbb{R}^{N \times D}$, and $\mathbf{cls} \in \mathbb{R}^D$.  

These tokens are fed into a global Transformer layer (\textcolor{darkyellow}{yellow} block in Fig.~\ref{fig:main}) to enhance contextual understanding of the global (WSI-level) information and improved feature representation :
\begin{equation}
    [\mathbf{cls}_1, \mathbf{P}_1, \mathbf{E}_1] = f_{\text{global}} ([\mathbf{cls}, \mathbf{P}_0, \mathbf{E}_0]),
\end{equation}
where $ [\mathbf{cls}_1, \mathbf{P}_1, \mathbf{E}_1]$ are the output of the global Transformer layer $f_{\text{global}}$. \\
\noindent\textbf{Prompt-based Clustering.}
With the refined prompt token embeddings $\mathbf{P}_1$ and $\mathbf{E}_1$, we dynamically group patches based on their distance to the $C$ prompt embeddings via projection. The previous token clustering methods in ViT for image tasks~\cite{marin2023token,zeng2022not} cluster tokens based on their pairwise similarities. However, these approaches become impractical in WSI settings, where the number of image tokens can exceed ten thousands in our cases, making the computation of pairwise distances prohibitively expensive due to high computational costs and resource constraints. To address this challenge, we introduce a novel approach leveraging learnable prompt tokens, each associated with a cluster as a proxy, for efficient clustering by projecting patch tokens onto the prompt tokens. 
We define an assignment matrix $\mathbf{A}\in \mathbb{R}^{N \times C}$, where each row corresponds to a patch token and each column to a cluster. The entry $\mathbf{A}^{c}_{i}$ indicates the probability that the $i$-th patch token belongs to cluster $c$ and is calculated follows:
\begin{equation}
    \mathbf{A}^c_{i} = \frac{e^{\left\langle \mathbf{E}^{i\cdot}_{1 }, \mathbf{P}^{c\cdot}_{1 }\right\rangle}}{\sum_{c'}^{C} e^{\left\langle \mathbf{E}^{i\cdot}_{1 }, \mathbf{P}^{c'\cdot}_{1 }  \right\rangle}},
    \label{eq:assign}
\end{equation}
where $\left\langle \cdot,\cdot \right\rangle$ is the inner product, $\mathbf{E}_1$ is the output of token feature after the global Transformer layer's self-attention module. 

\noindent\textbf{Prompt Updating.} 
Xavier uniform initialization \cite{glorot2010understanding} is used to randomly initialize $C$ uniform prompt tokens $\mathbf{P}_0 \in \mathbb{R}^{C\times D}$ to prevent clustering collapse. Then, the Gram-Schmidt process ensures orthogonality for $i \in [C]$:$
    \mathbf{u}_i = \mathbf{X}_{i\cdot} - \sum_{j=1}^{i-1}\frac{\left\langle \mathbf{u}_j, \mathbf{X}_{i\cdot} \right\rangle}{\left\langle \mathbf{u}_j, \mathbf{u}_j \right\rangle} \mathbf{u}_j, \mathbf{P}_{0i\cdot} = \frac{\mathbf{u}_i}{\| \mathbf{u}_i \|_2},
$
where $\mathbf{X} = [\mathbf{X}_1, \cdots, \mathbf{X}_C]$ is a set of randomly initialized vectors with Xavier uniform initialization and
$\mathbf{u} = [\mathbf{u}_1, \cdots, \mathbf{u}_C]$ is a set of orthonormal vectors.
Furthermore, we introduce a soft constraint as the regularization loss function to prevent prompt collapse across clusters during training, in addition to updates based on the downstream task loss (see Sec.~\ref{sec:pooling}). Explicitly, we minimize the difference between $\mathbf{P}_1^T \mathbf{P}_1$ and the identity matrix $\mathbf{I}$:
\begin{equation}
\mathcal{L}_{\text{reg}} = \| \mathbf{P}_1^{T}\mathbf{P}_1 - \mathbf{I}\|_2.
\end{equation} 
In the MIL problem, the batch size is typically set to $1$ due to memory limits to process vast amount patches in each WSI. To facilitate stable prompt updating, we utilize moving average strategy to update prompts: 
$\bar{\mathbf{P}}_{1m} = \theta\bar{\mathbf{P}}_{1{(m-1)}} + (1-\theta) \mathbf{P}_{1m},$
where $m$ is the number of steps in one epoch, $\theta \in [0,1]$ is the decay factor that controls how fast the prompts are updated, and $\bar{\mathbf{P}}$ indicate the averaged prompt over iterations. This approach smooths out prompt updates across batches, reducing sensitivity to individual slide differences.

\subsection{Merging to Obtain Prototypes over Clusters}
Next, we aim to learn the prototypes in each cluster. 
According to the assignment matrix obtained in Eq.~\eqref{eq:assign}, we have cluster index vector $\mathbf{a} = \arg \max (\mathbf{A}_{i\cdot}) = [a_1, a_2, \ldots, a_N]^T$, for $i \in [C]$.
With $\mathbf{a}$, we conduct cluster-wise re-index to patch tokens and get $\mathbf{H} = [\mathbf{H}_1, \mathbf{H}_2,\ldots,\mathbf{H}_C]$, where $\mathbf{H}_i$ is the concatenation of features in cluster $i$. The tokens in each cluster are another local Transformer layer (denoted as $f_{\text{local}}$ with output dimension $d$, shown as \textcolor{purple}{purple} blocks in Fig.~\ref{fig:main}, parameters shared for efficiency) to learn cluster-wise local context:
\begin{equation}
    [\mathbf{\bar{p}}^{c}_{2}, \mathbf{H}^{c}_{2}] = f_{\text{local}}([\mathbf{\bar{p}}^{c}_{1},\mathbf{H}^{c}_{1}]), \quad c=1,2,\ldots,C.
\end{equation}
where $\mathbf{\bar{p}}^{c}_{2} \in \mathbb{R}^{d}, \mathbf{H}^{c}_{2}\in\mathbb{R}^{N_i \times d}$ are the output from the $c$th Transformer for cluster $i$, which contains $N^c$ patches. 

To reduce redundancy before passing it to the pooling module, clustering-based MIL methods summarize representative information from each cluster as \emph{prototypes}. In clustering task, centroid of the data in the cluster is commonly used as prototype that provide good approximation of the entire cluster~\cite{jain1988algorithms}. Although we introduce learned prompts as proxies for clusters to enable efficient, learnable clustering, using $\mathbf{\bar{p}}_{2}$ as a candidate to represent clusters may deviate significantly from the actual cluster centers in practice (as illustrated in Fig.~\ref{fig:main}(b)). To address this, we propose to calculate the centroid among token embeddings $\mathbf{H}^{c}_{2}$ via merging to represent prototypes (represented as the \textcolor{orange}{orange} block in Fig.~\ref{fig:main}). Additionally, following \cite{rao2021dynamicvit}, we introduce learnable weights $\mathbf{r}_i = [r_1,\ldots,r_{N^c}]^T \in \mathbb{R}^{N^c}$, for $c \in [C]$,  to explicitly represent the averaging weights of the patch token features. Hence, the prototype $\mathbf{h}^c_{\text{P}}$ for cluster $i$ is written as via weighted averaging:
$\mathbf{h}^c_{\text{P}} = \frac{\sum_{j \in [N^c]} e^{r_j}\mathbf{h}^j_{2}}{\sum_{j \in [N^c]} e^{r_j}} \in \mathbb{R}^{d}$,
and we have $\mathbf{H}_\text{P} = [\mathbf{h}^1_{\text{P}}, \cdots, \mathbf{h}^C_{\text{P}}]$.
\subsection{Global Pooling for Downstream Tasks}
\label{sec:pooling}
Focusing on the most common WSI analysis tasks: classification and survival analysis, we detail the pooling module for these downstream applications. \\
\noindent\textbf{Classification.}
Due to the heterogeneity of contents in WSI, it is not suitable to only use a single class token to summarize the information over the whole WSI \cite{chen2022scaling}. With the help of prototypes, we can get better slide-level representation, $\mathbf{H}_{\text{final}} = [\mathbf{cls}_1, \mathbf{H}_\text{P}]$, and the final prediction of WSI is $\hat{Y} = \text{Pooling}( \mathbf{H}_{\text{final}})$. Specifically, the pooling here includes mean operation over $\mathbf{H}_{\text{final}}$ and a linear layer. To this end, the overall objective function for classification is:
\begin{align}
\mathcal{L}= \mathcal{L}_{\text{cla}} + \alpha \mathcal{L}_{\text{reg}}
 =\text{CE}(\hat{Y},Y)+ \alpha\|\bar{\mathbf{P}}^T\bar{\mathbf{P}}-\mathbf{I}\|_2,
 \label{eq:cls}
\end{align}
where $\hat{Y}$ is the prediction of WSI, $\alpha$ is the  regularization loss term weight.\\
\noindent\textbf{Survival Analysis.}
The survival analysis model is used to predict the survival hazard score, which can be formulated by
$f_{\text {hazard }}(T=t)=\lim\limits_{\partial t \rightarrow 0} \frac{P(t \leq T \leq t+\partial t \mid T \geq t)}{\partial t}$ $=\lambda_0(t) e^{\beta  \mathbf{H}_{\text{final}}}$,
which measure the probability of patient death instantaneously at $t$, and $\beta$ is the parameters of the last linear prediction layer. 
In training with WSIs, we follow \cite{Chen_2021_ICCV} to construct weak supervision of transforming the continuous observation time to discrete time intervals: $T_j=r, \text { if } T_{j, \text { cont }} \in\left[t_r, t_{r+1}\right) \text { for } r \in\{0,1,2,3\}$, where $j$ is the patient index, $T_{j, \text { cont }}$ is the continuous event time. For a given patient with bag-level feature $ \mathbf{H}_{\text{final}}$, the hazard function can be defined as:
$
f_{\text {hazard }}\left(r \mid \mathbf{H}_{\text{final}_j}\right)=P\left(T_j=r \mid T_j \geq r, \mathbf{H}_{\text{final}_j}\right)
$
and the survival function is 
$f_{\text {surv}}\left(r \mid \mathbf{H}_{\text {final}_j}\right)  =P\left(T_j > r \mid \mathbf{H}_{\text {final}_j}\right) 
 =\prod_{u=1}^r\left(1-f_{\text {hazard }}\left(u \mid \mathbf{h}_{\text {final }_j}\right)\right).$ 
The loss is log likelihood function of survival is:
\begin{align}
L_{\text{surv}} &= - c_j \log \left( f_{\text{surv}}(Y_j \mid \mathbf{H}_{\text{final}, j}) \right) \nonumber 
- (1 - c_j) \log \left( f_{\text{surv}}(Y_j - 1 \mid \mathbf{H}_{\text{final}, j}) \right) \nonumber \\
&\quad - (1 - c_j) \log \left( f_{\text{hazard}}(Y_j \mid \mathbf{H}_{\text{final}, j}) \right),
\label{eq:6}
\end{align}
where $c_j$ is the binary censorship status, $c_j = 1$ means the patient live longer than the follow-up period, $c_j = 0$ means the patient passed away within time $T_j$. Similar to Eq~\eqref{eq:cls}, we add $\mathcal{L}_\text{reg}$ to $\mathcal{L}_{\text {surv}}$ as the total loss for survival prediction. 


\section{Experiment}
\label{sec:experiment}
\subsection{Dataset}
\noindent\textbf{Classification.}
We evaluate PTCMIL on Camelyon16 (2-class) \cite{bejnordi2017diagnostic}, TCGA-Non-Small Cell Lung Cancer (NSCLC) (2-class) \cite{cancer2013cancer}, Prostate cANcer graDe Assessment (PANDA) (6-class) \cite{bulten2022artificial} and an in-house prostate WSI dataset (1-class). Camelyon16 is for detecting metastases (abnormal) (129) or normal (270) in breast cancer, TCGA-NSCLC is for subtyping the subtypes LUAD (538) and LUSC (512) of lung cancer. and PANDA is for grading Prostate cancer diagnosis (10,616). For Camelyon16 and TCGA-NSCLC, we follow \cite{lu2021data} to split the training, validation and testing sets and use five-fold validation to report the result. For PANDA, we use the data splits of the challenge. To evaluate the adaptability of our method, we use the in-house prostate WSI dataset (749 cancerous slides) for testing with the model trained on PANDA. 

\noindent\textbf{Survival Analysis.}
We evaluate the survival prediction performance on Breast Invasive Carcinoma (BRCA) (1,041), Colon and Rectum Adenocarcinoma (CRC) (575), Bladder Urothelial Carcinoma (BLCA) (437) and Lung adenocarcinoma (LUAD) (519) from \cite{cancer2013cancer}. We follow \cite{Song_2024_CVPR} to use 5-fold site-stratified cross-validation. 
\subsection{Implementation and Evaluation}
We use CTransPath \cite{wang2022transformer} and UNI \cite{chen2024towards} to extract patch feature with CLAM's toolbox \cite{lu2021data} to crop non-overlapping 256 $\times$ 256\ (20$\times$) patches.
We use Adam for our model and keep the original optimizers for baselines. 
Cosine scheduler is used with a starting learning rate of 2e-4. 
The weight decay is set to 1e-5. The regularization loss weight $\alpha$ is 0.1 for Camelyon16 and PANDA, 0.2 for all TCGA datasets, decay factor $\theta$ is 0.9 for all. The numbers of clusters for Camelyon16, TCGA and PANDA are 7, 5 and 5 respectively. 
For the cancer normal/abnormal and subtype classification tasks, we report accuracy and AUC, presenting mean and standard deviation. 
For PANDA, we report Cohen's kappa.
For the in-house dataset that contains only one class, we use accuracy. 
We report the concordance index (c-index) for survival prediction.

\subsection{Comparison with Baselines}
\noindent\textbf{Classification and Survival Analysis.}
Tables~\ref{tab:main}  shows the result ofclassification on Camelyon16 (abnormal detection), TCGA-NSCLC (subtyping), PANDA (grading), and in-house prostate (adaptation) datasets. We also conduct survival prediction analysis on four TCGA datasets in Table~\ref{tab:survival}. PTCMIL consistently demonstrates high performance, highlighting the end-to-end integration of clustering enables optimal WSI representation learning for various downstream task. 
\begin{table*}[!t]
\centering
\caption{Classification result on four datasets. {\footnotesize ($\dag$: We use the reported result from the original paper.)} } 
\label{tab:main}
\resizebox{0.95\textwidth}{!}{%
\begin{tabular}{@{}c|c|cc|cc|c|c@{}}
\toprule
\multirow{2}{*}{\makecell{Feature\\extraction}} & \multirow{2}{*}{Method} & \multicolumn{2}{c|}{Camelyon16} & \multicolumn{2}{c|}{TCGA-NSCLC} & \multicolumn{1}{c|}{PANDA} & \multicolumn{1}{c}{\makecell{PANDA $\to$ \\ in-house \\ prostate dataset}} \\ \cline{3-8} 
\multicolumn{1}{c|}{} & \multicolumn{1}{c|}{} & AUC & Acc & AUC & Acc & Cohen's $\kappa$ & Acc \\ 
\midrule
\multirow{10}{*}{CTransPath} 
& \multicolumn{1}{l|}{ABMIL (NeurIPS'18) \cite{ilse2018attention}} & $92.40_{4.17}$ & $90.31_{1.80}$ & $95.61_{1.88}$ & $89.81_{2.60}$ & $0.892$ & $85.81$ \\ 
& \multicolumn{1}{l|}{DSMIL (CVPR '21) \cite{li2021dual}} & $93.26_{2.83}$ & $87.03_{1.18}$ & $96.79_{0.94}$ & $90.87_{2.02}$ & $0.900$ &  $87.28$ \\ 
& \multicolumn{1}{l|}{CLAM (Nat. Biomed. Eng. '21) \cite{lu2021data}} & $95.89_{2.48}$ & $92.19_{1.91}$ & $97.13_{0.83}$ & $91.60_{1.36}$ & $0.915$  & $86.75$ \\ 
& \multicolumn{1}{l|}{DTFD-MIL (CVPR'22) \cite{zhang2022dtfd}} &$94.93_{1.32}$&$92.81_{3.09}$&$97.24_{0.43}$&$91.02_{1.72}$&$0.913$&$87.00$ \\ 
& \multicolumn{1}{l|}{TransMIL (NeurIPS'22) \cite{shao2021transmil}} & {$96.47_{1.12}$} & {$93.13_{2.56}$} & $96.67_{0.87}$ & $90.72_{0.74}$ & $0.897$ & $84.34$ \\ 
& \multicolumn{1}{l|}{ILRA (ICLR'23) \cite{xiang2023exploring}} & $94.29_{2.82}$ & $90.78_{2.02}$ & $96.33_{0.67}$ & $90.19_{1.07}$ & \bm{$0.928$} &  $84.07$ \\ 
& \multicolumn{1}{l|}{PANTHER (CVPR'24) \cite{Song_2024_CVPR}} & $67.01_{4.79}$ & $64.19_{6.01}$ & $93.39_{0.88}$ & $91.64_{2.30}$ & $0.720$ & $81.52$ \\ 
& \multicolumn{1}{l|}{MambaMIL(MICCAI'24) \cite{yang2024mambamil}} & $92.31_{1.37}$ & $91.09_{1.31}$ & $96.85_{1.10}$ & {$91.85_{0.69}$} & $0.902$ & $87.15$ \\ 
& \multicolumn{1}{l|}{DGR-MIL (ECCV'24) \cite{zhu2024dgr}} &$91.25_{7.18}$&$90.06_{3.60}$&$96.13_{1.17}$&$89.51_{1.86}$&$0.894$&$87.82$ \\ 
& \multicolumn{1}{l|}{PTCMIL (ours)} & \bm{$98.06_{0.90}$} & \bm{$94.73_{1.33}$} & \bm{$97.31_{0.67}$} & \bm{$92.17_{1.80}$} &  \bm{$0.928$} & \bm{$89.96$} \\ 
\midrule
\multirow{10}{*}{UNI} 
& \multicolumn{1}{l|}{ABMIL (NeurIPS'18) \cite{ilse2018attention}} & $98.42_{0.67}$ & $95.73_{2.92}$ & $97.72_{0.55}$ & $92.30_{1.55}$ & $0.935$ & $84.74$ \\ 
& \multicolumn{1}{l|}{DSMIL (CVPR '21) \cite{li2021dual}} & $98.75_{1.08}$ & $97.50_{1.02}$ & $97.56_{0.69}$ & $93.43_{1.30}$ & $0.857$ & $84.61$ \\ 
& \multicolumn{1}{l|}{CLAM (Nat. Biomed. Eng. '21) \cite{lu2021data}} & $98.92_{0.82}$ & {$97.97_{0.43}$} & $98.11_{0.46}$ & $93.21_{0.75}$ & $0.933$ &  $86.75$ \\ 
& \multicolumn{1}{l|}{DTFD-MIL (CVPR'22) \cite{zhang2022dtfd}} &$98.38_{0.74}$&$96.56_{2.38}$&$97.89_{0.55}$&$92.23_{1.66}$&$0.911$&$84.74$\\ 
& \multicolumn{1}{l|}{TransMIL (NeurIPS'22) \cite{shao2021transmil}} & {$99.08_{0.74}$} & $95.31_{3.95}$ & {$98.20_{0.30}$} & {$93.58_{0.80}$} & {$0.936$} & {$89.56$} \\ 
& \multicolumn{1}{l|}{ILRA (ICLR'23) \cite{xiang2023exploring}} & $94.38_{5.48}$ & $93.44_{4.51}$ & $96.72_{0.71}$ & $90.04_{1.64}$ & $0.924$ & $89.29$ \\ 
& \multicolumn{1}{l|}{PANTHER (CVPR'24) \cite{Song_2024_CVPR}} & $84.21_{6.02}$ & $79.19_{6.02}$ & $97.82_{0.67}$ & $91.92_{1.66}$ & $0.923^\dag$ & $87.42$ \\ 
& \multicolumn{1}{l|}{MambaMIL (MICCAI'24) \cite{yang2024mambamil}} & $99.06_{0.77}$ & $97.97_{1.96}$ & $97.96_{0.97}$ & $92.68_{1.27}$ & $0.929$ &$86.61$ \\ 
& \multicolumn{1}{l|}{DGR-MIL (ECCV'24) \cite{zhu2024dgr}} &$98.54_{2.03}$&$97.35_{0.89}$&$97.54_{0.52}$&$92.30_{1.57}$&$0.915$&$89.69$\\ 
& \multicolumn{1}{l|}{PTCMIL (ours)} & \bm{$99.60_{0.34}$} & \bm{$98.60_{0.35}$} & \bm{$98.44_{0.39}$} & \bm{$93.81_{1.02}$} & \bm{$0.937$} & \bm{$92.64$} \\ 
\bottomrule
\end{tabular}%
}
\end{table*}

\noindent\textbf{Adaptability.}
We explore few-shot (20 random WSIs with balanced labels) domain adaptation using limited WSIs to transfer learned prompt tokens to new tasks. We only update the classifier and prototypes (if have). Table~\ref{tab:adaptation} shows that a small number of prompt tokens enable effective cross-domain adaptation. This is promising for resource-constrained scenarios with limited training data, highlighting PTCMIL's adaptability and robustness across varied domains.

\subsection{Visualization and Interpretation}
\label{sec:visualization}
\begin{figure*}[t!]
  \centering
  \centerline{\includegraphics[width=\columnwidth]{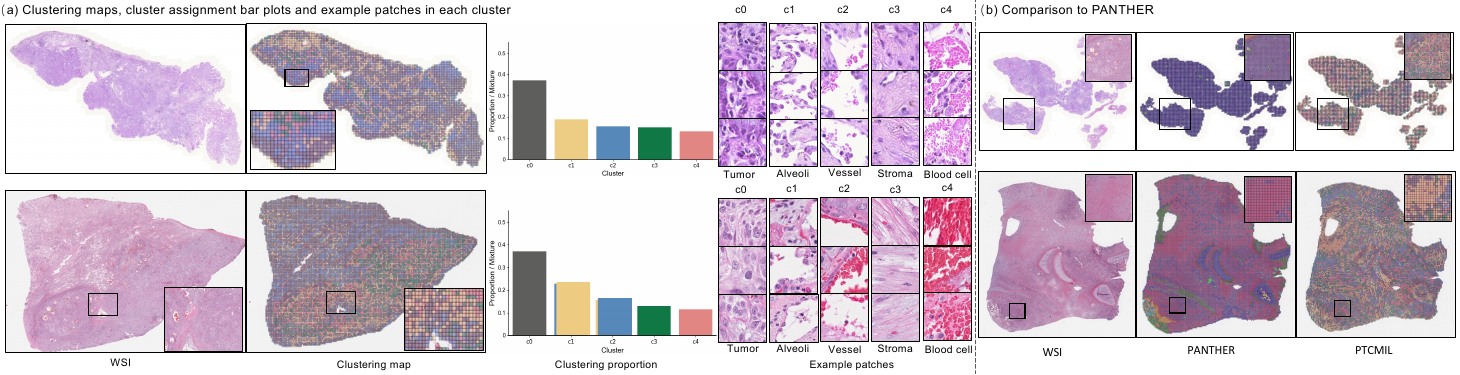}}
\caption{ Visualization and interpretation of PTCMIL. (a) Clustering maps, cluster assignment bar plots and example patches in each cluster; (b) Comparison to PANTHER. 
}
\label{fig:main_visual}
\end{figure*}

Fig.~\ref{fig:main_visual} shows (a) WSI clustering maps, cluster assignment bar plots and example patches in each cluster (b) comparison to PANTHER \cite{song2024morphological}. In details, most patches in c0 are related to tumor cells that are irregular in shape, with enlarged, darkly stained nuclei and often disordered arrangement, showing high mitotic activity. c1 are mainly tumor cells, which are characterized by irregular shapes, enlarged nuclei, disordered arrangements. Lung alveoli (c2) are thin-walled sacs primarily lined by flattened type I cells and cuboidal type II cells. The stroma (c3) consists of spindle-shaped cells within a collagen-rich extracellular matrix. c4 are mainly pools of red blood cells that appear as tightly packed, uniform red cells. Besides, in Fig.~\ref{fig:main_visual}(b), the two-stage clustering MIL model PANTHER \cite{song2024morphological} shows clustering collapse (homogeneous colors, poor tissue separation), while PTCMIL produces more structured maps, better reflecting local heterogeneity. 
\begin{table}[t]
    \centering
    \begin{minipage}[t]{0.45\textwidth}
        \centering
        \caption{Survival analysis (c-index).}
        \label{tab:survival}
        \resizebox{\textwidth}{!}{%
            \begin{tabular}{c|c|c|c|c}
                \toprule
                Method & LUAD & BLCA & BRCA & CRC \\
                \midrule
                DSMIL \cite{li2021dual} & $0.659_{0.07}$ & $0.586_{0.06}$ & $0.720_{0.06}$ & $0.696_{0.11}$ \\
                CLAM \cite{lu2021data} & $0.625_{0.12}$ & $0.603_{0.06}$ & $0.698_{0.03}$ & $0.678_{0.09}$ \\
                DTFD-MIL \cite{zhang2022dtfd} &$0.637_{0.08}$&$0.609_{0.08}$&$0.693_{0.05}$&$0.697_{0.09}$ \\
                TransMIL \cite{shao2021transmil} & $0.660_{0.12}$ & $0.616_{0.08}$ & $0.708_{0.05}$ & $0.686_{0.06}$ \\
                ILRA \cite{xiang2023exploring} & $\bm{0.688_{0.06}}$ & $0.603_{0.04}$ & $0.726_{0.08}$ & $0.704_{0.09}$ \\
                PANTHER \cite{Song_2024_CVPR} & $0.632_{0.07}$ & $0.612_{0.07}$ & $0.729_{0.08}$ & $0.632_{0.14}$ \\
                MambaMIL \cite{yang2024mambamil} & $0.670_{0.08}$ & $0.606_{0.04}$ & $0.668_{0.05}$ & $0.680_{0.06}$ \\
                DGR-MIL \cite{zhu2024dgr} &$0.674_{0.05}$&$0.608_{0.04}$&$0.658_{0.05}$& $0.700_{0.09}$\\
                PTCMIL (ours) & $\bm{0.688_{0.09}}$ & $\bm{0.630_{0.05}}$ & $\bm{0.745_{0.04}}$ & $\bm{0.738_{0.09}}$ \\
                \bottomrule
            \end{tabular}
        }
    \end{minipage}
    \hfill
    \begin{minipage}[t]{0.43\textwidth}
        \centering
        \caption{Few-shot adaptation (\%).}
        \label{tab:adaptation}
        \resizebox{\textwidth}{!}{%
            \begin{tabular}{c|cc|cc}
                \toprule
                Pretrained on & \multicolumn{2}{c|}{TCGA-NSCLC} & \multicolumn{2}{c}{Camelyon16} \\
                Fewshot on & \multicolumn{2}{c|}{Camelyon16} & \multicolumn{2}{c}{TCGA-NSCLC} \\
                \hline
                & AUC & Acc & AUC & Acc \\
                \midrule
                DSMIL \cite{li2021dual} & $65.57_{8.81}$ & $62.97_{6.76}$ & $83.88_{6.74}$ & $75.47_{6.15}$ \\
                CLAM \cite{lu2021data} & $62.04_{10.63}$ & $61.41_{4.12}$ & $84.72_{5.36}$ & $76.08_{5.15}$ \\
                DTFD-MIL \cite{zhang2022dtfd} &$66.47_{5.89}$&$62.50_{5.21}$&$83.51_{6.61}$&$66.87_{8.70}$\\
                TransMIL \cite{shao2021transmil} & $57.28_{6.14}$ & $53.44_{5.78}$ & $66.00_{10.59}$ & $61.36_{9.54}$ \\
                ILRA \cite{wang2023iteratively}&$50.15_{12.59}$&$53.91_{7.35}$&$53.73_{10.79}$&$53.46_{6.96}$\\
                MambaMIL \cite{yang2024mambamil} & $65.45_{10.14}$ & $65.31_{7.17}$ & $83.93_{5.71}$ & $75.62_{4.45}$ \\
                 DGR-MIL \cite{zhu2024dgr}&$55.84_{4.20}$&$62.03_{1.05}$&$54.00_{3.41}$&$51.09_{2.77}$ \\
                PTCMIL (ours) & $\bm{69.49_{10.27}}$ & $\bm{67.03_{10.53}}$ & $\bm{85.73_{3.42}}$ & $\bm{77.36_{3.89}}$ \\
                \bottomrule
            \end{tabular}
        }
    \end{minipage}
\end{table}

\subsection{Ablation Studies and Hyperparameter Analysis}
\begin{figure}[t!]
    \centering
    \begin{minipage}{0.52\textwidth} 
        \centering
        \begin{subfigure}[b]{0.3\linewidth}
            \includegraphics[width=\linewidth]{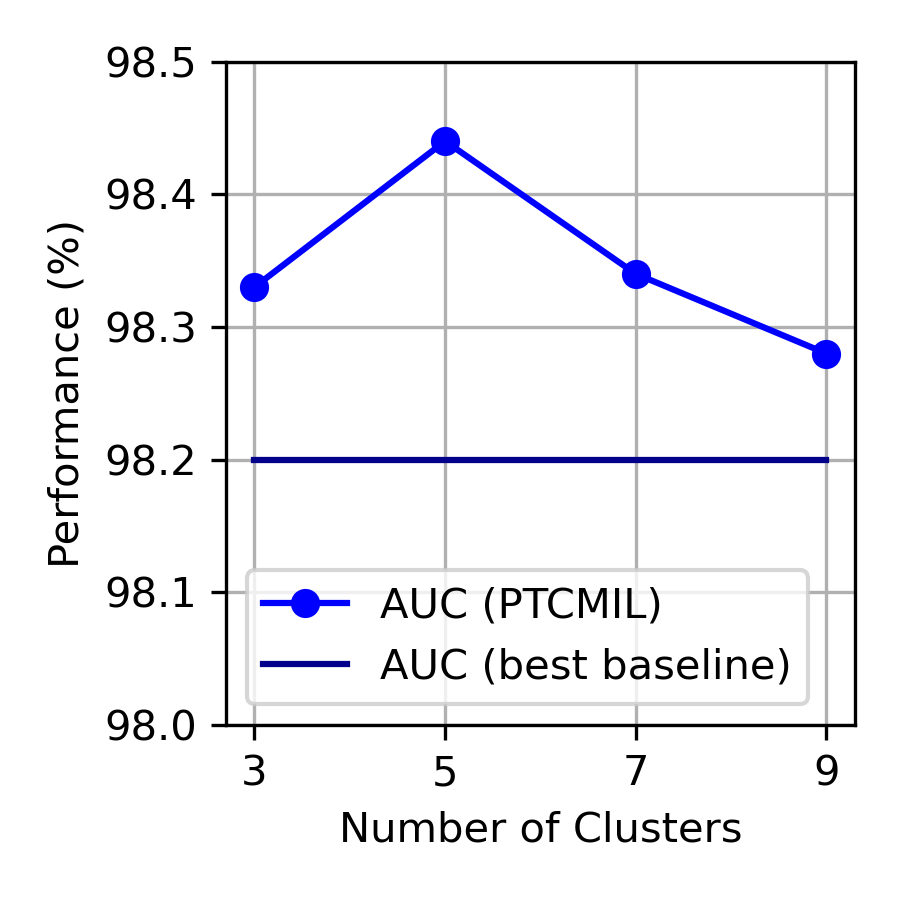}
            \caption{NSCLC}
            \label{fig:clu_nsclc}
        \end{subfigure}
        \hfill
        \begin{subfigure}[b]{0.3\linewidth}
            \includegraphics[width=\linewidth]{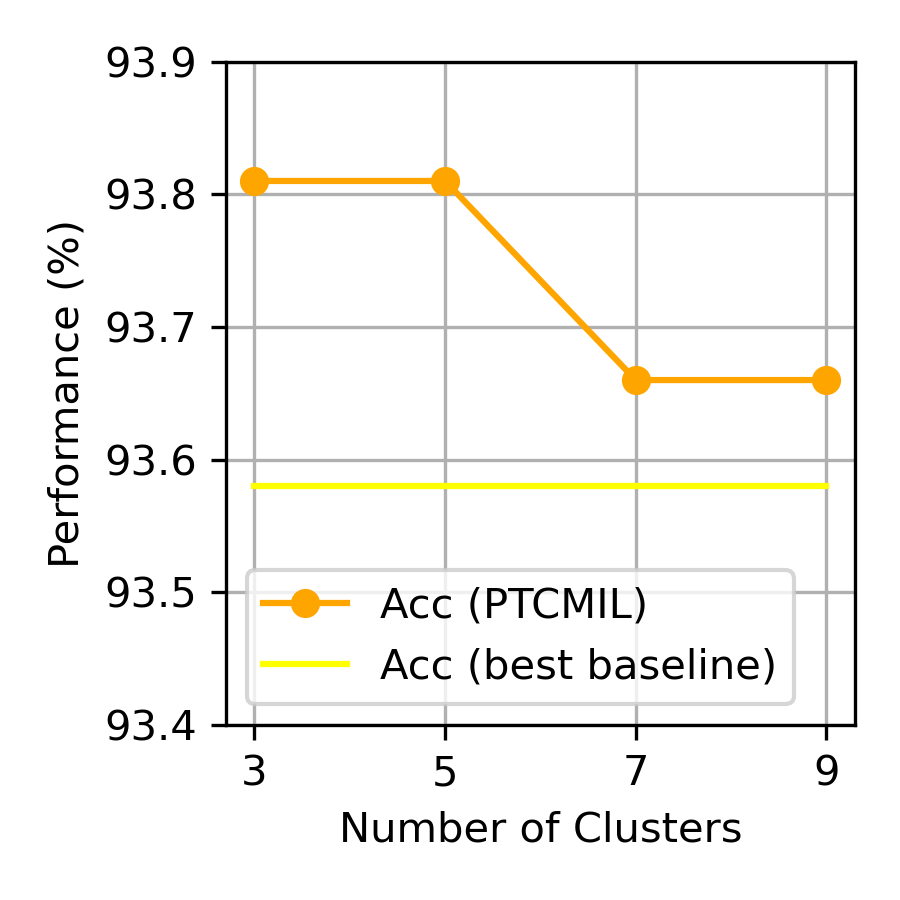}
            \caption{NSCLC}
            \label{fig:clu_crc}
        \end{subfigure}
        \hfill
        \begin{subfigure}[b]{0.3\linewidth}
            \includegraphics[width=\linewidth]{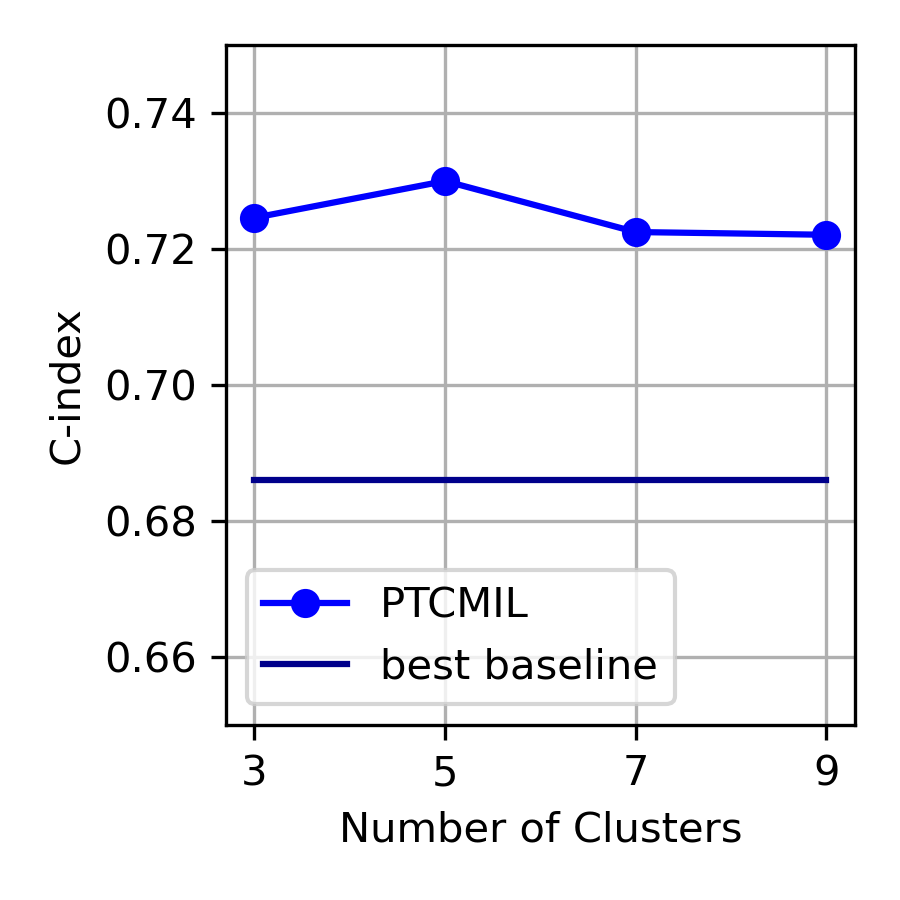}
            \caption{CRC}
            \label{fig:clu_crc}
        \end{subfigure}
        
\        \caption{Variation in the number of clusters. }
        \label{fig:cluster_num}
    \end{minipage}
    \hfill
    \begin{minipage}{0.47\textwidth} 
        \centering
        \captionof{table}{Ablation of clustering, merging, and pooling.}
        
        \label{tab:ablation}
        \resizebox{\linewidth}{!}{%
        \begin{tabular}{cccc|ccc|cc|c}
        \hline
        \multicolumn{2}{c|}{Clustering} & \multicolumn{2}{c|}{Merging} & \multicolumn{3}{c|}{Pooling} & \multicolumn{2}{c|}{TCGA-NSCLC} & TCGA-CRC \\ \hline
        \multirow{2}{*}{w/o} & \multicolumn{1}{c|}{\multirow{2}{*}{w/}} & \multirow{2}{*}{w/o} & \multirow{2}{*}{w/} & \multirow{2}{*}{pro} & \multirow{2}{*}{cls} & \multirow{2}{*}{pro + cls} & \multirow{2}{*}{AUC} & \multirow{2}{*}{Acc} & \multirow{2}{*}{c-index} \\ 
        & \multicolumn{1}{c|}{} &  &  &  &  &  &  &  &  \\ \hline
        $\checkmark$ & \multicolumn{1}{c|}{} & $-$  & $-$  & $-$ & $-$ & $-$ & $96.58_{0.45}$ & $90.87_{0.68}$ & $0.705_{0.07}$ \\ \hline
        & \multicolumn{1}{c|}{$\checkmark$} & $\checkmark$ &  &  &  & $\checkmark$ & $96.77_{0.79}$ & $94_{1.87}$ & $0.685_{0.08}$ \\ \hline
        & \multicolumn{1}{c|}{$\checkmark$} &  & $\checkmark$ & $\checkmark$ &  &  & $96.92_{0.72}$ & $91.77_{1.01}$ & $0.726_{0.07}$ \\
        & \multicolumn{1}{c|}{$\checkmark$} &  & $\checkmark$ &  & $\checkmark$ &  & $96.44_{0.54}$ & $95_{0.87}$ & $0.706_{0.07}$ \\
        \rowcolor{yellow} & \multicolumn{1}{c|}{$\checkmark$} &  & $\checkmark$ &  & &  $\checkmark$ & \bm{$97.31_{0.67}$} & \bm{$92.17_{1.80}$} & \bm{$0.738_{0.09}$} \\ \hline
        \end{tabular}%
        }
    \end{minipage}
    
\end{figure}

We present ablation studies on key modules in Table~\ref{tab:ablation}. \textbf{Clustering (sec 1):} We evaluate the effectiveness of clustering in ViT-based MIL aggregation and its sensitivity to the number of clusters. PTCMIL (last line) achieves higher AUC and accuracy for classification on TCGA-NSCLC and improved survival prediction on CRC. \textbf{Merging (sec 2):} We assess the impact of merging tokens to create prototypes versus directly using prompt tokens. Merging reduces token redundancy and enhances WSI representation, leading to better performance. In contrast, using prompt tokens alone results in lower performance, as further illustrated in Fig.~\ref{fig:main}(c). \textbf{Pooling (sec 3):} 
While the cls token provides global image representation in ViT, integrating prototype tokens alongside the {cls} token enhances performance, 
showing the advantage of prototype-guided pooling. 
In Fig.~\ref{fig:cluster_num}, PTCMIL consistently outperforms the best baseline within a cluster range of 3 to 9, achieving peaks at number of cluster equals 5 on TCGA-NSCLC (classification) and CRC (survival). 

\section{Conclusion}
\label{sec:conclusion}
In this paper, we propose PTCMIL, an end-to-end clustering ViT-based MIL for WSI feature aggregation, addressing WSI's giga-pixel scale and heterogeneity. By introducing learnable prompt tokens and integrating clustering with prediction, PTCMIL handles WSI heterogeneity effectively. Experiments show superior performance across tasks and improved slide clustering where prior methods struggled. Our approach enables simultaneous prototype learning and task performance enhancement while identifying interpretable biomarkers. Future work will explore automatic cluster number selection for cancer types and the integration of vision-language models with clinical knowledge for guided clustering.

\begin{credits}

\subsubsection{\discintname}
The authors have no competing interests to declare that are
relevant to the content of this article.
\end{credits}

{
\bibliographystyle{splncs04}
\bibliography{main}
}


\end{document}